# Chest X-ray Pneumothorax Segmentation Using EfficientNet-B4 Transfer Learning in a U-Net Architecture

Alvaro Aranibar Roque

*Freie Universität Berlin, Germany*

*Pontificia Universidad Católica del Perú, Perú*

aranibar.a@pucp.edu.pe

Helga Sebastian

*Freie Universität Berlin, Germany*

helgas93@zedat.fu-berlin.de

*Abstract— Pneumothorax, the abnormal accumulation of air in the pleural space, can be life-threatening if undetected. Chest X-rays are the first-line diagnostic tool, but small cases may be subtle. We propose an automated deep-learning pipeline using a U-Net with an EfficientNet-B4 encoder to segment pneumothorax regions. Trained on the SIIM-ACR dataset with data augmentation and a combined binary cross-entropy plus Dice loss, the model achieved an IoU of 0.7008 and Dice score of 0.8241 on the independent PTX-498 dataset. These results demonstrate that the model can accurately localize pneumothoraces and support radiologists.*

*Keywords— U-net, CNN, Deep learning, Pneumothorax, Segmentation, Transfer learning*

## I. INTRODUCTION

Pneumothorax is the abnormal accumulation of air in the pleural space, which can arise spontaneously or due to trauma or medical procedures. Early detection is critical, as even small pneumothoraces may rapidly progress to life-threatening conditions. Clinical examination alone may miss subtle cases [1], making chest X-rays the standard diagnostic tool. On frontal radiographs, pneumothorax appears as a sharply demarcated radiolucent region along the lung edge, but small or low-contrast cases are often difficult to identify. In such cases, higher-sensitivity imaging like CT or ultrasound may be required for confirmation [2].

Automated segmentation using deep learning has emerged as a promising solution for detecting subtle pneumothoraces. Encoder–decoder networks such as U-Net preserve both high-level context and fine spatial details, while EfficientNet backbones provide scalable, high-quality feature extraction. Large annotated datasets, including the SIIM-ACR Pneumothorax Challenge, have enabled the development of accurate CNN-based segmentation pipelines. Building on these advances, our study employs a U-Net with an EfficientNet-B4 encoder, trained on the SIIM-ACR dataset with data augmentation and a combined Dice–cross-entropy loss. The model is evaluated on the independent PTX-498 dataset.

## II. BACKGROUND

### A. Pneumothorax and X-ray imaging

Pneumothorax is a medical condition in which air abnormally accumulates in the pleural space between the lung and the chest wall, causing partial or complete lung collapse [3]. It can occur spontaneously (e.g. rupture of subpleural blebs) or secondary to trauma or invasive procedures, and is potentially life-threatening if not recognized promptly. In practice, small pneumothoraces may be difficult to detect on clinical exam (e.g. diminished breath sounds, hyperresonant percussion), so imaging is crucial for diagnosis [2]. The standard diagnostic test is a chest X-ray (radiograph), because the contrast between air and lung tissue is readily visible on radiographs. In fact, large-scale datasets for pneumothorax segmentation (e.g. the 2019 SIIM-ACR Pneumothorax Challenge [4]) comprise only posteroanterior chest X-rays annotated for pneumothorax regions.

X-ray radiography uses an external X-ray source that passes ionizing radiation through the patient to a detector, producing a 2D image of internal structures. Tissues attenuate (absorb or scatter) X-rays according to their density: high-density structures (such as bone) block more X-rays and appear bright (white), whereas low-density regions (such as air in the lungs) transmit X-rays and appear dark (black) [2]. For example, the normally aerated lung fields appear dark gray/black on a chest radiograph. Chest X-rays are widely available, fast, inexpensive, and non-invasive [5], making them the first-line imaging study for suspected pneumothorax. On a frontal chest X-ray, a pneumothorax typically appears as a sharp, radiolucent (black) region along the lung edge, with absence of vascular markings in the affected area. If a plain X-ray is equivocal, higher-sensitivity modalities like computed tomography (CT) or ultrasound may be used for confirmation [2].

*B. U-Net Architecture*

U-Net [6] is a convolutional neural network with a U-shaped design, originally proposed for biomedical image segmentation. As shown in Fig. 1, it consists of two symmetric paths: an encoder that progressively compresses the input image while extracting hierarchical features, and a decoder that reconstructs the spatial resolution to produce a dense segmentation map. The encoder applies blocks of two 3×3 convolutions with ReLU activations followed by 2×2 max-pooling, reducing the spatial resolution while doubling the number of feature channels at each step. The decoder employs 2×2 transposed convolutions (upsampling) to recover spatial resolution, halving the number of channels, and concatenates these feature maps with the corresponding encoder maps through skip connections, followed by additional 3×3 convolutions. Being fully convolutional (without fully connected layers), U-Net directly outputs a pixel-wise segmentation map, allowing efficient processing of input images of arbitrary size.

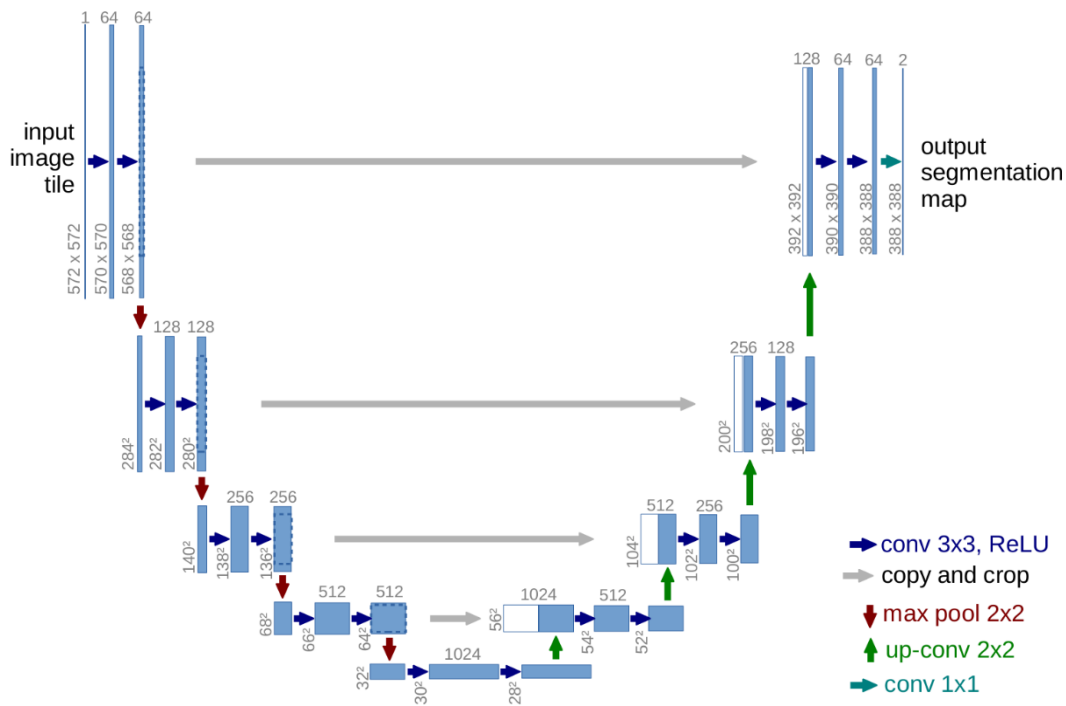

Fig. 1. U-Net Architecture [6].

A key component of U-Net is the use of skip connections, which link encoder and decoder layers at the same spatial scale [6]. These connections transfer high-resolution feature maps from the encoder to the decoder, preserving fine spatial details (e.g., edges, textures) that would otherwise be lost due to pooling [6]. As a result, the decoder integrates both global contextual information and fine-grained local details, enabling more precise reconstruction of object boundaries.

The encoder–decoder fully convolutional design makes U-Net especially well-suited for medical image segmentation. Since the model processes the entire image in a single forward pass, it outputs a dense, pixel-level classification that assigns each pixel to its respective category (organ, background, lesion, etc.) [6]. Moreover, U-Net was designed to cope with limited annotated datasets: its training relies heavily on data augmentation strategies (rotations, elastic deformations, translations, among others) to enhance generalization with relatively few labeled images.

In the case of pneumothorax, these architectural properties are particularly advantageous. On chest X-rays or CT scans, the visual manifestation of pneumothorax is often limited to a faint pleural line with very low contrast [7]. U-Net's skip connections retain the low-level image information required to detect such subtle boundaries, while the encoder provides contextual anatomical information to guide accurate delineation.

## C. EfficientNet backbone.

The width, depth, and input resolution of a CNN are hyperparameters that were often adjusted arbitrarily. EfficientNet is a family of convolutional network architectures designed by Tan and Le [8] using a compound scaling method. Increasing the model's complexity typically leads to higher accuracy but also requires greater computational capacity. While accuracy may initially improve significantly as model complexity increases, it eventually reaches a saturation point where accuracy no longer improves, and only computational cost continues to grow. As shown in Fig. 1, when parameters are scaled independently, the accuracy saturation point reaches at most around 79.5%. However, the study demonstrated that there is a dependency among scaling dimensions. Instead of arbitrarily scaling parameters of a baseline network, EfficientNet uniformly scales all dimensions (depth, width, and input resolution) with a compound coefficient in a principled way [8] because there exists a correlation between parameters complexity. As shown in Fig. 1, compound scaling can achieve higher accuracy. This systematic scaling yields models (EfficientNet-B0 through B7) that achieve a favorable trade-off between accuracy and efficiency.

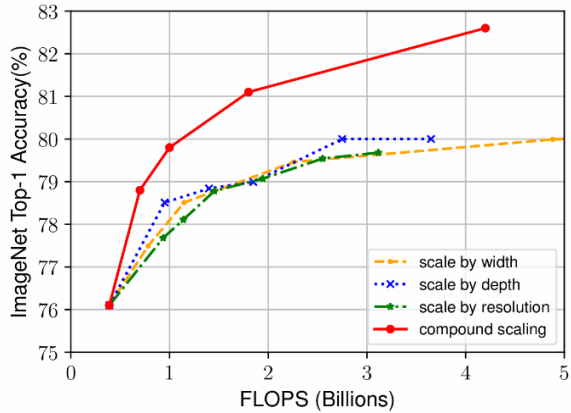

Fig. 2. Scaling Up EfficientNet-B0 with Different Methods [8].

Notably, the largest model (EfficientNet-B7) reached state-of-the-art ImageNet accuracy (84.3% top-1) while being many times smaller and faster than previous networks [8]. EfficientNet models also transfer well to other tasks, making them attractive backbones for segmentation.

## III. RELATED WORK

In 2019, the SIIM-ACR Pneumothorax Segmentation Challenge [9] was hosted on Kaggle, attracting over 1,400 teams in the development phase and more than 300 in the final stage. The leading solutions consistently relied on deep convolutional neural networks, particularly U-Net and its variants with pre-trained encoders. The winning team achieved first place using U-Net with ResNet-34, ResNet-50, and SE-ResNeXt-50 backbones, applying a progressive training strategy with increasing image resolution, strong augmentation, and tailored post-processing [2]. The second-place team employed a two-stage approach, separating classification from segmentation, and leveraged an ensemble of U-Net and DeepLabv3 models with EfficientNet and SE-ResNeXt encoders. Other top teams introduced innovations such as attention mechanisms (e.g., CBAM), deep supervision, and semi-supervised training with external datasets to further enhance generalization. Table 2, taken from [2], shows the solutions for the top 5 teams in the challenge.

TABLE 1
SOLUTIONS FOR THE TOP 5 TEAMS IN THE PNEUMOTHORAX CHALLENGE [2].

| Rank | Team | Network | Encoder | Techniques | Score |
|---|---|---|---|---|---|
| 1 | [dsmlkz] sneddy | U-Net | ResNet (34, 50), SE-ResNext-50 | Triplet threshold | 0.8679 |
| 2 | X5 | Deeplabv3+, U-Net | SE-ResNext (50, 101) EfficientNet (B3, B5) | Segmentation with Classification | 0.8665 |
| 3 | Bestfitting | U-Net | ResNet-34, SE-ResNext-50 | Lung segmentation and CBAM attention | 0.8651 |
| 4 | [ods.ai] amirassov | U-Net | ResNet-34 | Deep supervision | 0.8644 |
| 5 | Earhian | U-Net | SE-ResNext (50, 101) | ASPP and Semi-supervision | 0.8643 |

Subsequent studies have extended these approaches, experimenting with multi-resolution strategies, semi-supervision, and staged training pipelines. For instance, the 2ST-UNet method introduced a two-phase training scheme where networks were first trained at lower resolution before fine-tuning on high-resolution images, demonstrating improved convergence and stability. Across these works, common practices included transfer learning with ImageNet-pretrained encoders, aggressive data augmentation, and test-time augmentation (TTA) combined with ensemble strategies.

## IV. METHOD

### A. Dataset description

This study employed two independent datasets. On one side, for training and internal validation, the X-ray dataset provided by the 2019 SIIM-ACR Pneumothorax Segmentation Challenge [9] was used. Since, at the time of writing this paper, the dataset was no longer available in the Cloud Healthcare API from the original source, a copy shared on Kaggle [4] was used. The dataset consists of 12,047 images and their corresponding masks. On the other side, for the test and measurement of the final metrics, the PTX-498 dataset presented by Wang et al. [10] was used, which consists of 498 images and their respective masks. The latter is divided into three folders, as the data were collected from three hospitals in Shanghai and labeled by two senior radiologists.

The images of the original Segmentation Challenge dataset were in Digital Imaging and Communications in Medicine (DICOM) format. These depicted chest X-rays of patients. Therefore, the images are 2D, even though the standard can store 3D images in the form of slices. Furthermore, they may contain metadata with additional information such as age, imaging technique, patient position, and other data that could be useful as inputs for a more complex architecture that goes beyond the scope of this paper. However, in the copy of the dataset we worked with and in the second dataset PTX-498, the images are already converted into PNG format with a resolution of 1024x1024 and 3 channels.

In the original Segmentation Challenge dataset, the masks were encoded in the lossless compression format RLE. However, in the acquired copy, the masks were already decoded into 1024x1024 PNG images with values of 255 (white pixel) in the pneumothorax sections and 0 (black pixel) in the non-pneumothorax sections. The masks of PTX-498 are also in the same format.

Table 2 shows an overview of the distributions of the datasets. The Segmentation Challenge dataset was imbalanced since it contains 2,669 images with pneumothorax. This was randomly divided into 85% training set and 15% validation set. Finally, after training the model, the PTX-498 dataset will be used to measure the final metrics. It is important to mention that the latter only contains X-rays with positive pneumothorax. The reason for its use and its implications will be discussed in the following sections.

TABLE 2
DATASET OVERVIEW

| Attribute | Segmentation Challenge | PTX-498 |
|---|---|---|
| **Number of cases** | 12047 | 498 |
| **Number of positive cases** | 2669 | 498 |
| **Number of negative cases** | 9378 | 0 |

### B. Data pre-processing and augmentation

In the top solutions of the Pneumothorax Segmentation Challenge, different image resolutions were used for training neural networks, including 1024x1024, 768x768, 512x512, and 256x256 [2]. Due to hardware limitations and in order to reduce computational load, GPU memory usage, and training time, lower resolutions were often adopted. For example, in [2], it is shown that the training time per epoch ranged from 10.3 minutes to 23.3 minutes depending on the architecture used, the number of parameters, the batch size and the GPU computational power. Their training experiments were conducted on an NVIDIA Titan Xp GPU (12 GB memory) with 32 GB RAM. Additionally, downsized images of 512x512 from the Segmentation Challenge and batch sizes between 4, 6, and 60 were employed. Besides these parameters, the total training time also depends on the number of epochs. In the experiments reported in [2], between 45 and 100 epochs were used, which implied **training times between 14 h 36 min and 53 h 18 min**. Considering that the hardware and time available for our work are limited (further detailed in the Experiments section), and given that many of the winning solutions also adopted it, we chose to train with downsized images of 512x512. This resolution also represents a middle ground since, on the one hand, very high resolutions may hinder precise boundary localization and increase the risk of the model "memorizing" noise, thereby affecting generalization [11], while on the other hand, very low resolutions imply loss of fine details, textures, and subtle information.

The images and masks were stored in two separate folders. The standard *Python* modules *os* and *glob* were used to save the file paths in two string lists. These were then used to create a *tf.data* pipeline that loads the dataset into memory on demand (lazy loading). That is, the 12,047 images of the dataset are not loaded simultaneously, as there would not be enough memory to perform such an operation. Instead, image and mask tensors are loaded in batches during training. In [2], small batch sizes of 4 and 6 were used, considering that the GPU used for those experiments had only 12 GB of VRAM. In our case, taking advantage of an RTX 3090 GPU with 24 GB of VRAM and the use of Mixed Precision training, which reduces memory consumption, we were able to use a batch size of 32 and thus shorten training time.

Data augmentations in segmentation improve model generalization and reduce overfitting by artificially increasing the diversity of the training set, creating realistic variations of the original images that help the model learn more robust patterns. In addition to resizing to 512x512 pixels, Fig. 3 shows the augmentations applied to the dataset using the *albumentations* library, which provides higher performance compared to the default transformations in Keras and TorchVision [12]. Moreover, it can be seamlessly integrated into the *tf.data* pipeline, and unlike Keras, it natively allows simultaneous transformations of both images and masks.

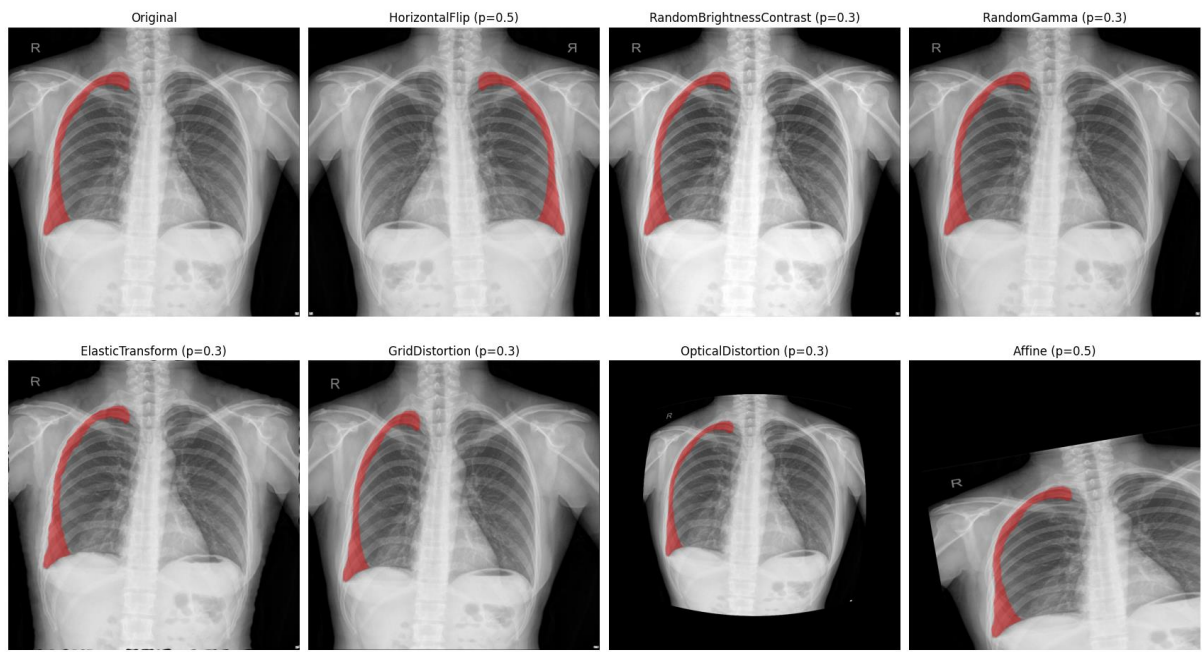

Fig. 3. Example of augmentations applied with their respective probabilities.

Two augmentation objects were created: *train_transform* and *val_transform*. The former includes the transformations, in addition to resizing, applied during training with their respective probabilities detailed in Fig. 3. The latter includes only the 512x512 resizing used during validation. In the *_load_and_augment* function, image tensors are loaded with the standard *cv2* module according to their file paths and the corresponding training or validation transformations are applied. This function is integrated into the pipeline through *tf.py_function*.

*C. Architecture and transfer learning*

Transfer learning refers to initializing a model from weights learned on a different (usually larger) dataset. In medical imaging, where labeled data are scarce, transfer learning is widely used to boost performance. A common strategy is to take a CNN pretrained on ImageNet and fine-tune it on the target task. The pre-trained encoder, known as the backbone, provides general-purpose visual features (edges, textures, etc.) which greatly accelerate convergence and improve accuracy on the medical dataset. This strategy will also help reduce training time, which is important in our study due to hardware limitations. In fact, top-performing pneumothorax segmentation pipelines typically use ImageNet-pretrained backbones.

In our work we adopt EfficientNet-B4 as the U-Net encoder implemented using the *segmentation_models* library from Keras. B4 is a mid-sized EfficientNet (between B0 and B7) that offers high representational power with moderate computational cost. In the literature, using EfficientNet-B4 as a pretrained encoder has proven effective: for instance, Liu et al. [13] reported that a U-Net with an EfficientNet-B4 encoder significantly improved lung segmentation accuracy on chest X-rays.

Since the task is binary segmentation (presence or absence of pneumothorax), the final layer uses a sigmoid activation to produce pixel-wise probabilities in the range [0, 1]. This activation is standard in binary segmentation tasks as it allows for probabilistic interpretation of each pixel prediction.

The loss function combines binary cross entropy (BCE) with Dice loss with equal weights as shown in (1) and proposed in [2]. As shown in (2), BCE penalizes pixel-wise misclassifications by assigning larger loss values to incorrect predictions with low predicted probability for the true class, as a consequence of its logarithmic formulation. As shown in (3), Dice Loss is a loss function derived from the Dice coefficient, also known as the F1-score in binary segmentation tasks. This coefficient quantifies the overlap between the predicted mask (A) and the reference mask (B), evaluating the degree of agreement between the region segmented by the model and the ground truth. Unlike pixel-wise error-based losses, Dice Loss emphasizes the overlap of positive areas where the region of interest often represents only a small fraction of the image.

$$Combined\ loss\ =\ H_q + DCSL \qquad (1)$$

$$H_q\ =\ -\frac{1}{N}\sum_i^N (1 - y_i) \cdot log(1 - p(y_i)) + y_i \cdot log(p(y_i)) \qquad (2)$$

$$DCSL = 1\ - \frac{2 \cdot |A \cap B|}{|A| + |B|} \qquad (3)$$

This combination is particularly effective for segmentation problems with severe class imbalance, such as pneumothorax detection, where affected regions can be very small relative to the entire image. BCE ensures precise pixel-level predictions, while Dice emphasizes capturing the target region, resulting in more reliable and coherent segmentations.

For optimization, we adopted Adam with a cosine decay learning rate schedule. This approach starts with a relatively high learning rate (e.g., 1e-3) and gradually decreases it following a cosine curve toward a lower bound (e.g., 1e-5) throughout training. Cosine annealing schedules, inspired by SGDR, have been shown to improve convergence by allowing rapid learning in early stages and fine-tuning during later stages [14]. Practically, this prevents the learning rate from remaining too high once the model approaches convergence, resulting in a more stable training process and improved segmentation accuracy.

*D. Post-processing*

The raw output of the segmentation network consists of probability maps, where each pixel encodes the likelihood of belonging to the pneumothorax region. In order to transform these continuous probability values into discrete binary masks, we applied a binarization threshold BT. Specifically, pixels with predicted probability values greater than BT were assigned to the pneumothorax class (value = 1), while the rest were assigned to the background (value = 0).

However, thresholding alone can result in the presence of small spurious regions, typically originating from model uncertainty or noise. To address this, we incorporated a secondary post-processing stage based on connected component analysis. In this step, all connected regions in the binary mask smaller than a removal threshold (RT) were eliminated, under the assumption that clinically meaningful pneumothorax regions exhibit a minimum area. This procedure was also used in the top winning solutions [2].

The optimal values of both BT and RT were not fixed a priori but determined through a grid search over the validation dataset. By systematically evaluating different threshold combinations, we identified the pair of values that maximized the Intersection over Union (IoU) between predictions and ground truth masks.

## V. EXPERIMENT

### A. Experiment setup

RunPod is a cloud platform designed for training neural networks, offering access to GPUs via Pods or Serverless environments, along with tools like JupyterLab or SSH for development. Users can train models by configuring GPU resources and connecting through code. For this study, a Pod was rented in Secure Cloud equipped with an **RTX 3090 GPU** featuring 24 GB of VRAM, 22 GB of RAM, and Ubuntu 22.04 preinstalled with Python 3.11, CUDA 12.8.1, and cuDNN. The hourly cost was **$0.46**, with an additional $0.008/hr for a Volume Disk used for data storage.

The development process employed standard Python libraries: *os, glob, random, math, json*, and *typing* for file management, mathematical operations, and data handling. Furthermore, external libraries such as Keras and Segmentation Models were used to define pretrained segmentation architectures, *Albumentations* for image augmentation, *OpenCV* for visual processing, *Matplotlib* for results visualization, and *Scikit-learn* for data partitioning and performance metrics evaluation. The use of the MONAI framework was considered for the workflow; however, it was ultimately not adopted, as its advantages are primarily oriented toward 3D DICOM images, in contrast to the 2D PNG images employed in this study.

Table 3 summarizes the parameters employed. In this case, due to the available GPU VRAM and to reduce training time, a batch size of 32 was used, compared to the batch size of 4 adopted in [2], which had access to 12 GB of VRAM. The implications of this decision will be discussed in the following section. To compensate for the reduced number of updates per epoch, training was conducted for 300 epochs, in contrast to the 80 epochs reported in [2]. In addition, an early stopping call-back was implemented, halting training if the validation IoU did not improve for 20 consecutive epochs. Furthermore, mixed precision was employed to reduce training time, which consists of using lower-precision (float16) operations where possible, while maintaining higher-precision (float32) calculations for critical steps.

TABLE 3
TRAINING PARAMETERS

| U-Net Backbone | EfficientNetB4 |
|---|---|
| Optimizer | Adam with cosine decay |
| LR Schedule | 1e−4 to 1e−6 |
| Batch Size | 32 |
| Epochs | 300 |

### B. Evaluation with PTX-498 dataset

To evaluate segmentation quality, we employed overlap-based metrics, namely Intersection over Union (4) and F1-score (Dice coefficient), which capture agreement between predicted and ground-truth masks in terms of precision and recall. Internal validation was performed at the end of each epoch on the full validation set, monitoring IoU and F1 alongside the training loss (Dice + Binary Cross-Entropy). The IoU score was used for model checkpointing and early stopping.

$$IoU = \frac{|Ground\ truth\ area \cap Predicted\ area|}{|Ground\ truth\ area \cup Predicted\ area|} \quad (4)$$

After training, the validation was conducted on the PTX-498 independent dataset, where post-processed predictions were assessed using IoU, F1, Accuracy, Precision, Recall, and the confusion matrix to provide a comprehensive performance characterization. Accuracy reflects the overall proportion of correctly classified pixels but may be misleading due to class imbalance, as the pneumothorax region is typically small compared to the background. Recall (sensitivity) measures the ability to detect all pneumothorax pixels, critical to avoid missing true cases, while precision quantifies how many of the pixels predicted as pneumothorax are correct, thus reducing false alarms. The F1-score balances recall and precision, providing a robust measure of both detection sensitivity and specificity. Finally, the Intersection over Union (IoU) evaluates the spatial overlap between predicted and reference masks, serving as a direct indicator of segmentation accuracy in delineating pneumothorax regions.

## VI. RESULTS AND DISCUSSION

Table 4 presents a summary of the time measurements and the metrics obtained with the independent validation set PTX-498. The train time/epoch is lower than in [2] because the batch size is larger, taking advantage of the higher GPU VRAM as mentioned earlier.

TABLE 4
TIME MEASUREMENTS AND VALIDATION METRICS IN PTX-498

| | |
|---|---|
| **Train step time (ms)** | 450 ms |
| **Train time/epoch** | 2 m 24 s |
| **Total training time** | 12 h 15 min |
| **Accuracy** | 98.42% |
| **Recall** | 74.18% |
| **Precision** | 92.69% |
| **F1-score** | 82.41% |
| **IoU** | 70.08% |

Fig. 4 shows the resulting confusion matrix. Additionally, the optimal binary threshold was found to be 0.05, and the removal threshold was 0.

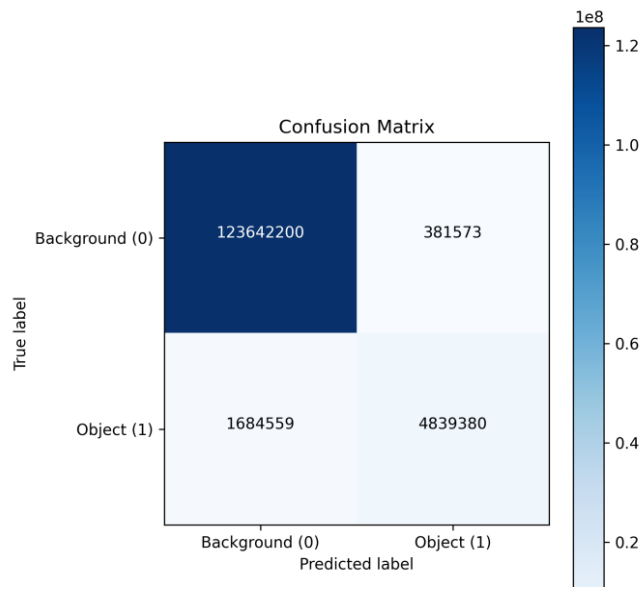

Fig. 4. Confusion matrix after validation with PTX-498 dataset

For this study, an F1-score (Dice coefficient) of 82.41% was achieved compared to the 86.79% obtained by the first-place solution in the challenge [4]. The lower performance may be due to some of the transformations applied to our model, such as affine and Optical Distortion, which were too aggressive and caused the lungs and masks to appear displaced, incomplete, or deformed, as shown in Fig. 3. It is also worth noting that the PTX-498 validation set contains only positive chest X-ray images, meaning performance could decrease when negative diagnosis images are included as input. Nevertheless, this limitation could be addressed with a two-stage architecture, similar to the top-2 solution: a CNN for classification followed by another for segmentation, such that the second stage is fed only with positive cases [9].

On the other hand, while the chosen batch size of 32 reduces training time and larger batches allow for more stable gradient estimates, they often worsen generalization and require more extensive learning rate adjustments [15]. Conversely, small batch sizes such as 4–6, used in [2], introduce useful noise that acts as a regularizer, thereby improving generalization.

Although an algorithm was developed to plot IoU over the training epochs, a programming error prevented the IoU plots from being saved to disk. Nevertheless, the metrics were monitored throughout training. Similar to the results reported in [2], IoU increased significantly during the first epochs; however, after 100 epochs, the improvements occurred in much smaller steps.

Fig. 5 shows a comparison between the labeled masks and the predictions from our model for six random images taken from the PTX-498 dataset. It can be observed that in large pneumothorax regions, the predicted area substantially overlaps with the ground truth; however, smaller regions may be missed.

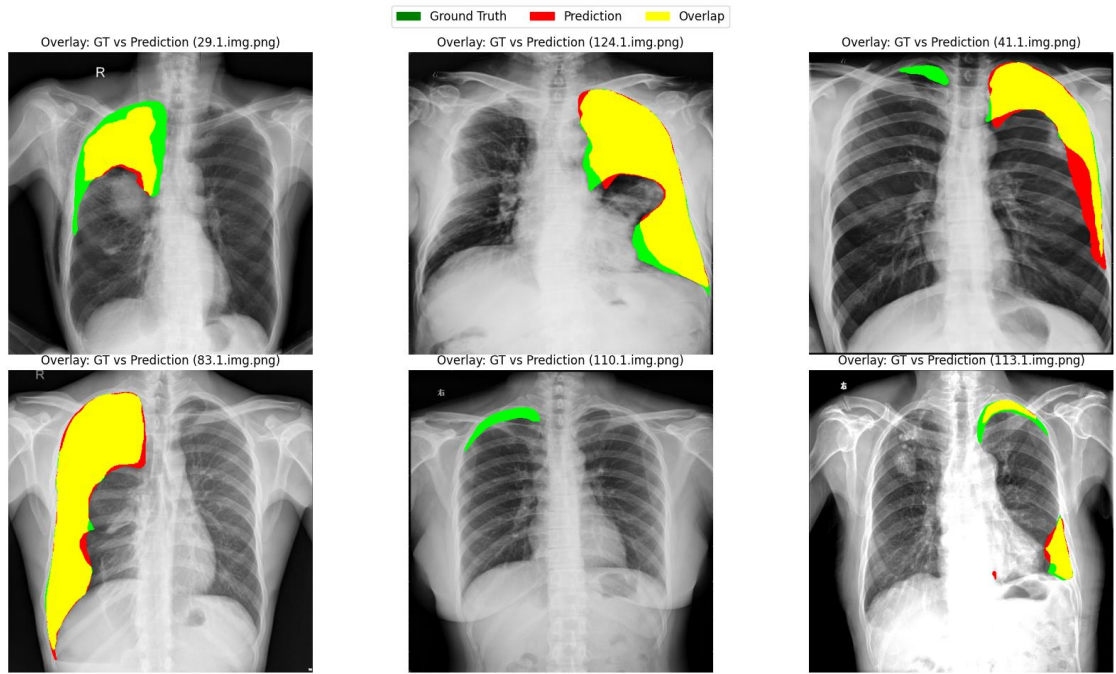

Fig. 5. Comparison of the predicted and labeled masks for six random images taken from the PTX-498 dataset.

## VII. Conclusions

We developed a U-Net with a pretrained EfficientNet-B4 encoder for automatic pneumothorax segmentation on chest X-rays. Post-processing steps, including probability thresholding and removal of small regions, refined predicted masks. Evaluation on the PTX-498 dataset yielded an IoU of 0.7008 and a Dice score of 0.8241, demonstrating that the model can localize large pneumothorax regions; however, smaller regions may be missed.

Although the F1 score is below the top SIIM-ACR challenge results, this difference may be due to aggressive augmentations and big batch sizes. Future improvements include integrating a preliminary classification stage, ensembling architectures, refining augmentation strategies, and employing multi-stage training to enhance boundary delineation. Overall, the results confirm that a U-Net with an EfficientNet backbone is an effective architecture for pneumothorax segmentation and could support radiologists in clinical decision-making.

## References

[1] Y.-H. Chan, Y.-Z. Zeng, H.-C. Wu, M.-C. Wu, and H.-M. Sun, "Effective Pneumothorax Detection for Chest X-Ray Images Using Local Binary Pattern and Support Vector Machine," *J Healthc Eng*, vol. 2018, p. 2908517, Apr. 2018, doi: 10.1155/2018/2908517.
[2] A. Abedalla, M. Abdullah, M. Al-Ayyoub, and E. Benkhelifa, "Chest X-ray pneumothorax segmentation using U-Net with EfficientNet and ResNet architectures," *PeerJ Comput Sci*, vol. 7, p. e607, June 2021, doi: 10.7717/peerj-cs.607.
[3] M. Noppen and T. De Keukeleire, "Pneumothorax," *Respiration*, vol. 76, no. 2, pp. 121–127, 2008, doi: 10.1159/000135932.
[4] "SIIM-ACR Pneumothorax Segmentation," vbookshelf. Accessed: Aug. 29, 2025. [Online]. Available: https://kaggle.com/siim-acr-pneumothorax-segmentation
[5] W. Liu, J. Luo, Y. Yang, W. Wang, J. Deng, and L. Yu, "Automatic lung segmentation in chest X-ray images using improved U-Net," *Sci Rep*, vol. 12, no. 1, p. 8649, May 2022, doi: 10.1038/s41598-022-12743-y.

[6] O. Ronneberger, P. Fischer, and T. Brox, "U-Net: Convolutional Networks for Biomedical Image Segmentation," May 18, 2015, *arXiv*: arXiv:1505.04597. doi: 10.48550/arXiv.1505.04597.

[7] J. I. S. Dumbrique, R. B. Hernandez, J. M. L. Cruz, R. M. Pagdanganan, and P. C. Naval, "Pneumothorax detection and segmentation from chest X-ray radiographs using a patch-based fully convolutional encoder-decoder network," *Front Radiol*, vol. 4, p. 1424065, Dec. 2024, doi: 10.3389/fradi.2024.1424065.

[8] M. Tan and Q. V. Le, "EfficientNet: Rethinking Model Scaling for Convolutional Neural Networks," Sept. 11, 2020, *arXiv*: arXiv:1905.11946. doi: 10.48550/arXiv.1905.11946.

[9] "2019 SIIM-ACR Pneumothorax Segmentation Challenge," Society for Imaging Informatics in Medicine (SIIM). Accessed: Aug. 29, 2025. [Online]. Available: https://kaggle.com/siim-acr-pneumothorax-segmentation

[10] Y. Wang *et al.*, "DeepSDM: Boundary-aware pneumothorax segmentation in chest X-ray images," *Neurocomputing*, vol. 454, pp. 201–211, Sept. 2021, doi: 10.1016/j.neucom.2021.05.029.

[11] J. P. Horwath, D. N. Zakharov, R. Mégret, and E. A. Stach, "Understanding important features of deep learning models for segmentation of high-resolution transmission electron microscopy images," *npj Comput Mater*, vol. 6, no. 1, p. 108, July 2020, doi: 10.1038/s41524-020-00363-x.

[12] "Albumentations: fast and flexible image augmentations," Albumentations. Accessed: Aug. 30, 2025. [Online]. Available: https://albumentations.ai/

[13] W. Liu, J. Luo, Y. Yang, W. Wang, J. Deng, and L. Yu, "Automatic lung segmentation in chest X-ray images using improved U-Net," *Sci Rep*, vol. 12, no. 1, p. 8649, May 2022, doi: 10.1038/s41598-022-12743-y.

[14] S. Jin, S. Yu, J. Peng, H. Wang, and Y. Zhao, "A novel medical image segmentation approach by using multi-branch segmentation network based on local and global information synchronous learning," *Sci Rep*, vol. 13, no. 1, p. 6762, Apr. 2023, doi: 10.1038/s41598-023-33357-y.

[15] D. Masters and C. Luschi, "Revisiting Small Batch Training for Deep Neural Networks," Apr. 20, 2018, *arXiv*: arXiv:1804.07612. doi: 10.48550/arXiv.1804.07612.